\documentclass[conference]{IEEEtran}
\IEEEoverridecommandlockouts
\usepackage{cite}
\usepackage{amsmath}
\usepackage{amssymb}
\usepackage{amsfonts}
\usepackage{algorithmic}
\usepackage{graphicx}
\usepackage{esvect}
\usepackage{stfloats}
\usepackage{threeparttable}
\usepackage{pifont}
\usepackage{multirow}
\usepackage{amssymb}
\usepackage{array}
\usepackage{amsmath}
\usepackage{textcomp}
\usepackage{xcolor}
\def\BibTeX{{\rm B\kern-.05em{\sc i\kern-.025em b}\kern-.08em
    T\kern-.1667em\lower.7ex\hbox{E}\kern-.125emX}}

\newcolumntype{P}[1]{>{\centering\arraybackslash}m{#1}}

\begin{document}

\title{On the Extreme Variance of Certified Local Robustness Across Model Seeds
}

\author{\IEEEauthorblockN{Minh Le}
\IEEEauthorblockA{\textit{Georgia Institute of Technology} \\
Atlanta, Georgia, United States \\
minhle@gatech.edu}
\and
\IEEEauthorblockN{Phuong Cao}
\IEEEauthorblockA{\textit{University of Illinois Urbana-Champaign} \\
Urbana, Illinois, United States \\
pcao3@illinois.edu}
}

\maketitle

\begin{abstract}
Robustness verification of neural networks, referring to formally proving that neural networks satisfy robustness properties, is of crucial importance in safety-critical applications, where model failures can result in loss of human life or million-dollar damages. However, the dependability of verification results may be questioned due to sources of randomness in machine learning, and although this has been widely investigated for accuracy, its impact on robustness verification remains unknown.

In this paper, we demonstrate a concerning result: Models that differ only in random seeds during training exhibit extreme variance in their certified robustness, with a standard deviation that is statistically larger than the marginal robustness improvements reported in recent machine learning papers. In addition, we also show that certified robustness generalization to unseen data varies significantly across datasets, falling short of the dependability expectations for safety-critical tasks.

Our findings are major concerns because: (i) machine learning results in certified robustness are likely unconvincing due to extreme variance in certified robustness, and (ii) a ``lucky'' model seed in a test set cannot be guaranteed to maintain its higher certified robustness under a different test set.

In light of these results, we urge researchers to increase the reporting of confidence intervals for certified robustness, and we urge those verifying neural networks to be more comprehensive in verification by using large-scale, diverse, and unseen data.
\end{abstract}

\begin{IEEEkeywords}
local robustness verification, safety-critical, random seeds, neural networks
\end{IEEEkeywords}

\maketitle

\begin{figure*}[b]
    \centering
    \includegraphics[width=\textwidth]{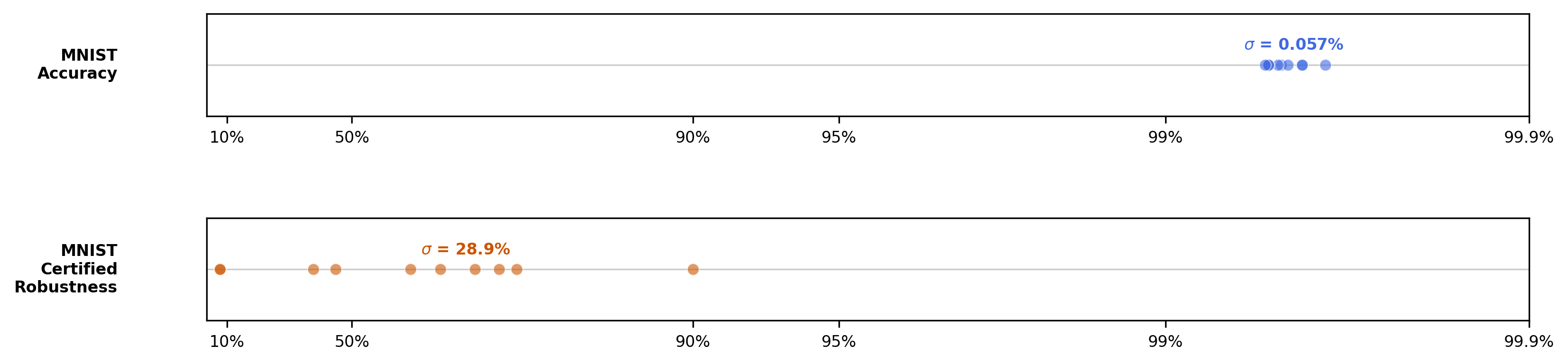}
    \caption{\textbf{Extreme variance in certified robustness for MNIST models.} Horizontal axis is plotted with a log scale. Models with 0.057\% stddev in accuracy (mean = 99.468\%) exhibit a \textbf{28.9\% stddev in certified robustness} (mean = 54.3\%) under perturbation level $\epsilon=0.007$.}
    \label{mnist_variance}
\end{figure*}

\section{Introduction}

\begin{table}[h!]
    \centering
    \begin{threeparttable}
        \begin{tabular}{cc|c}
            \hline
            \textbf{Paper} & \textbf{CI reported} & \textbf{Improvement}\\
            \hline
            Salman et al. (2019) \cite{salman2019provably}  & \ding{56} & 4.4\%\\
            Zhang et al. (2019) \cite{zhang2019towards} & \ding{56} & \textbf{9.85\%}\\
            Balunovic et al. (2020) \cite{balunovic2020adversarial} & \ding{56} & 6.5\%\\
            Zhang et al. (2021a) \cite{zhang2021towards} & \ding{56} & 2.04\% \\
            Zhang et al. (2021b) \cite{zhang2021boosting} & \ding{56} & 4.64\% \\
            Cullen et al. (2022) \cite{cullen2022double} & \ding{56} & 5\%\\
            Mueller et al. (2022) \cite{mueller2022certified} & \ding{56} & 7.26\%\\
            Hu et al. (2023) \cite{hu2023recipe} & \ding{56} & \underline{8.5\%}\\
            Altstidl et al. (2024) \cite{altstidl2024scalability} & \ding{56} & 3.95\%\\
            Lai et al. (2025) \cite{lai2025enhancing} & \ding{56} & 2.7\%\\
            Hu et al. (2026) \cite{hu2026lipnext} & \ding{56} & 8\%\\
            \hline
            \multicolumn{2}{c|}{\textbf{JPL certified robustness stddev (max)}} & \textbf{11.8\%}\\
            \multicolumn{2}{c|}{\textbf{MNIST certified robustness stddev (max)}} & \textbf{28.9\%}\\
            \hline
            \\
        \end{tabular}
        \caption{Stddev in certified robustness exceeding marginal improvements in 11 papers without CI report}
        \label{paper_variance}
        \begin{tablenotes}\footnotesize
            \item[1] Salman et al. (2019) uses both $l_2$ and $l_\infty$ distance for verification. The improvement reported here for this paper is for $l_\infty$ distance only since this paper only uses $l_\infty$ distance.
            \item[2] For improvements reported in papers, we \textbf{bold} the largest percentage point improvement and \underline{underline} the second largest.
        \end{tablenotes}
    \end{threeparttable}
\end{table}

Robustness verification of neural networks, referring to formally proving that neural networks satisfy robustness properties \cite{katz2017reluplex}, is of crucial importance in safety-critical applications, where model failures can result in loss of human life or million-dollar damages. This task is urgent because deep learning models are increasingly deployed in safety-critical domains, such as Tesla Autopilot in autonomous transportation \cite{Autopilo92:online}, Airborne Collision Avoidance System (ACAS) in aviation \cite{williams2004airborne}, PRO-AID from Argonne National Laboratory in nuclear reactor monitoring \cite{vilim2020advanced}, and FDA-approved medical devices \cite{Artifici20:online} from GE HealthCare \cite{GEHealth88:online}, Siemens Healthineers \cite{FDAClear40:online}, and Philips \cite{PhilipsD29:online}. However, much research on neural network robustness verification does not report any uncertainty quantification or confidence intervals, including those in mission-critical domains such as autonomous driving \cite{bernardeschi2025verifying}, medical imaging \cite{andreasenparallel}, aircraft lifespan \cite{kirov2023formal}, and power systems \cite{venzke2020verification}. This is a cause for concern because the performance of machine learning models can vary due to sources of randomness in training, and while many papers have explored this issue with respect to accuracy \cite{eryilmaz2024investigating} \cite{jordan2023variance} \cite{picard2021torch} \cite{bui2025assessing}, we are unaware of any similar work on certified robustness. In other words, the confidence in the models' reported certified robustness is unknown, falling well short of expectations for safety-critical tasks.

\textbf{This paper is among the first to demonstrate a concerning result for neural network verification: Models that only differ by random seeds during training have extreme variance in their certified robustness.} In particular, MNIST models with a 0.057\% standard deviation (stddev) in accuracy can exhibit a 28.9\% stddev in certified robustness at the right perturbation level. Most notably, \textbf{these standard deviations are statistically larger than marginal robustness improvements in recent machine learning papers} (see Table \ref{paper_variance}). We also test whether a ``lucky'' model seed on one test set can maintain its above-average certified robustness on another test set, and find that \textbf{certified robustness generalization varies significantly across datasets}. Although there is a strong correlation on MNIST, this correlation is weak on a Mars Frost Identification dataset from the NASA Jet Propulsion Laboratory \cite{MLreadyD65:online}, which represents a safety-critical task.

These findings are a major concern for several reasons. Firstly, machine learning results with certified robustness are now likely unconvincing. This is because the marginal improvements achieved are strictly less than the certified robustness standard deviation for all papers we survey, and are statistically smaller for all but one paper. Secondly, a ``lucky'' model seed in a test set cannot be guaranteed to maintain its higher certified robustness under a different test set. This means that attempts to circumvent this issue by simply selecting the most highly certified robust model seed are ill-founded and unlikely to succeed.

In light of these results, we urge researchers working on certified robustness to \textbf{report uncertainty quantification or confidence intervals} to be transparent about whether their models can be trusted in safety-critical applications. For those verifying neural networks, we urge them to use \textbf{large-scale, diverse, and unseen datasets, including those not directly used for training}, to more comprehensively evaluate whether the certified robustness of verified models is genuinely generalizable to unseen data.

\textbf{Approach.} The key steps in our work are:

\begin{itemize}
    \item \textit{Training:} 10 ResNet4 models (different seeds) trained on MNIST to 99.4\% test accuracy, and repeated on a Mars Frost Identification dataset with different parameters for a real-world safety-critical task.
    \item \textit{Verification}: Attempts to prove that no adversarial examples exist within an $L_\infty$ distance of $\epsilon$ for 100 inputs per model.
    \item \textit{Generalization:} Split the 100 inputs selected into two equal halves and measure the correlation in certified robustness between these two test sets.
\end{itemize}

\textbf{Major results.} Our paper consists of two major findings:

\begin{itemize}
    \item \textit{Neural networks with different seeds that achieve very similar accuracies have vastly different certified robustness.} In particular, MNIST models have a 0.057\% standard deviation (stddev) in accuracy, while getting 28.9\% stddev in certified robustness. Models trained on the Mars Frost Identification dataset have a 0.198\% stddev in accuracy while having 11.8\% stddev in certified robustness.
    \item \textit{Whether ``lucky'' model seeds maintain their certified robustness in a second test set is heavily dataset-dependent.} We find that the certified robustness of models, as evaluated on two separate test sets, is strongly correlated for MNIST (minimum correlation of 96.6\%) but only weakly on the Mars Frost Identification dataset (minimum correlation of 31.1\%), with a substantial difference in correlations.
\end{itemize}

\textbf{Future works.} This paper presents our preliminary work in demonstrating the extreme variance of certified robustness across different model seeds. Given that neural network verification is highly computationally expensive, we choose to verify a small number of model seeds on a single model architecture as a starting point, which we find is sufficient to demonstrate our claim. In future works, we will expand our experiments to include more random seeds, model sizes, model architectures (fully connected, convolutional), and safety-critical datasets (in e.g. aviation, healthcare, and autonomous driving).

\section{Related Works}

\begin{table}
    \centering
    \begin{tabular}{P{1.5cm}|P{3cm}|P{2.5cm}}
        \hline
        \textbf{Paper} & \textbf{Contribution} & \textbf{Our Difference}\\
        \hline
        \multicolumn{3}{P{8cm}}{\textbf{Group 1: Impact of Seeding on Accuracy}} \\
        \hline
        Picard \cite{picard2021torch} & Seed variance & \multirow{5}{2.5cm}{\centering Impact of seeding on \textit{certified robustness}, including for safety-critical task} \\
        \cline{1-2}
        Jordan \cite{jordan2023variance} & Seed generalization across test sets & \\
        \cline{1-2}
        Schader et al. \cite{schader2024don} & Seeding in safety-critical task (medicine)  & \\
        \hline
        \multicolumn{3}{P{8cm}}{\textbf{Group 2: Random Sources for Robustness Verification}} \\
        \hline
        Jia \& Rinard \cite{jia2021exploiting} & Floating-point error & \multirow{5}{2.5cm}{\centering \textit{Random seed} as another source of randomness} \\
        \cline{1-2}
        Manino et al. \cite{manino2025floating} & Floating-point error benchmark & \\
        \cline{1-2}
        VNN-COMP \cite{kaulen20256th} & Hardware differences & \\
        \hline
    \end{tabular}
    \vspace{5mm}
    \caption{Comparison of our paper with related works}
    \label{related_works}
\end{table}

\subsection{Impact of Random Seeds}

Much existing research has investigated the impact of seeding on the variance in neural network performance. Picard (2021) \cite{picard2021torch} shows that ResNet50 models trained on ImageNet have a 0.1\% stddev in accuracy and a 0.5\% gap between the maximum and minimum accuracies across only 50 seeds, arguing that this is a concerning result, given that a 0.5\% difference in ImageNet accuracy is widely considered significant. Our work arrives at the same conclusion for certified robustness, but with a significantly more extreme stddev (28.8\% vs. 0.1\%) and maximum-minimum gap (83\% vs. 0.5\%) from even fewer chosen seeds (10 vs. 50). Another paper by Jordan (2023) \cite{jordan2023variance} demonstrates, among other results, that random seeds performing better in one set of test data ``performing no better on average'' on a second set, with a correlation of only 3.1\%.

Our paper performs similar experiments on certified robustness, and although we obtained much higher correlations (96.3\% \& 45.5\% vs. 3.1\%), the significant variation in correlations across datasets means we cannot reliably conclude that certified robustness reliably generalizes across test sets.

\subsection{Random Sources for Robustness Verification}

Existing research shows that sources of randomness, such as floating-point errors and varying hardware, can lead to incorrect verification results. Jia \& Rinard (2021) \cite{jia2021exploiting} present an algorithm to efficiently find counterexamples to incorrect robustness claims and a method to construct neural networks that exploit incomplete verifiers. Manino et al. (2025) \cite{manino2025floating} develop a benchmark for neural network verification targeting floating-point errors, where 3\% of the verdicts of verification tools are incorrect. In addition, the latest Verification of Neural Networks Competition (VNN-COMP 2025) finds that running the same verification on different hardware can give different results, also due to floating-point errors \cite{kaulen20256th}.

Our paper complements existing works by investigating another common source of non-determinism in machine learning, namely random seeds, and shows that it also substantially reduces confidence in certified robustness claims due to extreme robustness variance and varying test set generalization. Together with existing literature, we demonstrate that non-determinism is a major unsolved challenge for neural network verification to be dependable and secure.

\section{Background}

\subsection{Local Robustness Verification}

Local robustness verification of a classification neural network refers to proving that, for a given input, there does not exist an adversarial input within the $L_p$ ball of radius $\epsilon$ centered at the original input, where the model gives a different output \cite{singh2018fast}. Formally, let the neural network be $f: \mathbb{R}^n \rightarrow \mathbb{R}^m$, where $n$ is the input dimension and $m$ is the number of classes. Consider an input $\mathbf{x} \in \mathbb{R}^n$ to the neural network. We wish to prove that:
$$\forall \mathbf{x_0} \in \mathbb{R}^n, \|\mathbf{x_0}-\mathbf{x}\|_p \leq \epsilon \Rightarrow \text{argmax}(f(\mathbf{x_0})) = \text{argmax}(f(\mathbf{x}))$$

Essentially, local robustness verification guarantees that the neural network will not change its output when the input is subjected to a (possibly adversarial) small perturbation. This is important in real-world tasks because input data can be noisy for various reasons, such as sensor noise \cite{rodriguez2024impact} \cite{geirhos2018generalisation} or uncleaned data \cite{sambasivan2021everyone} \cite{li2021cleanml}. The verification component is crucial in safety-critical tasks because we want some level of assurance that the model will not fail; otherwise, it can endanger human lives or cause major financial losses.

In this paper, we use the $L_\infty$ norm, following other research on local robustness verification \cite{zhang2022branch}, neural network verifiers, and 

\subsection{Neural Network Verifiers and $\alpha, \beta$-CROWN}

Neural network verifiers are used to verify the robustness of neural networks. There has been a wide range of verifiers, such as Marabou \cite{katz2019marabou}, Neurify \cite{wang2018formal}, VeriNet \cite{henriksen2020efficient}, nnenum \cite{bak2020improved}, and NeuralSAT \cite{duong2025neuralsat}, to name a few.

Currently, the state-of-the-art verifier is $\alpha, \beta$-CROWN \cite{zhou2025clip}, which has won the Verification of Neural Networks Competition from 2021 to 2025 \cite{kaulen20256th}. The main techniques of $\alpha, \beta$-CROWN include GPU-accelerated linear bound propagation \cite{xu2021fast} and branch-and-bound optimization \cite{wang2021beta}. Latest developments to $\alpha, \beta$-CROWN involve Clip-and-Verify \cite{zhou2025clip}, which tightens intermediate bounds during brand-and-bounding. 

In this paper, we use $\alpha, \beta$-CROWN for neural network verification, given its clear superiority over other verifiers in the past few years.

\section{Methodology}

\subsection{Datasets}

We use two datasets as our benchmark for training neural networks: the MNIST dataset \cite{lecun2010mnist} and a Mars Frost Identification dataset published by the NASA Jet Propulsion Laboratory (JPL) \cite{MLreadyD65:online}.

We include the MNIST dataset because it is a commonly used benchmark in neural network verification research \cite{koenig2024critically} \cite{zhang2022branch}. Given that neural network verification is a challenging and computationally expensive task, MNIST is often used as a starting point for evaluating neural network verifiers.

We also include the Mars Frost Identification dataset by JPL \cite{MLreadyD65:online} to demonstrate our results in a mission-critical setting. The dataset is used for binary classification of whether an image of the Mars surface contains frost. This task is important because future Mars missions rely on identifying habitable environments \cite{SUDSChal72:online}. Existing work on Mars' frost cycle focuses either on large-scale, low-resolution maps or on detailed maps of a small site \cite{diniega2025holistic}. A machine learning approach to frost identification will enable large-scale, high-resolution frost maps for future Mars exploration.

We perform standard data preprocessing on the dataset, where we normalize the data and randomly rotate images in the training partition only by up to 10 degrees to improve generalization. For simplicity, we also convert the images in the JPL dataset to MNIST format, i.e., convert them to grayscale and downscale to 28 $\times$ 28.

\subsection{Training of Neural Networks}

We choose a ResNet model architecture with 4 layers because experiments show it is sufficiently complex to achieve reasonable accuracy on both datasets while remaining small enough for easier robustness verification. 

For each dataset, we train 10 neural networks using seeds from 10 to 100 in increments of 10. We use cross-entropy loss and the AdamW optimizer with a weight decay of $10^{-4}$. We also use a learning rate scheduler with an initial learning rate of $10^{-4}$ that is multiplied by 0.3 every 30 epochs. 

We stopped training once the test set accuracy reached a threshold selected through experimentation. This threshold is 99.4\% for the MNIST dataset and 84\% for the JPL dataset, enabling fairer comparisons of the models' certified local robustness. 

\subsection{Robustness Verification of Neural Networks}

For each neural network, we verify its local robustness properties against 100 inputs from its corresponding dataset, using the $\alpha, \beta$-CROWN verifier. These inputs are the first 100 inputs from the test partition only to avoid data leakage. We perform verification for a range of perturbations $\epsilon$, where the mean certified robustness is high for small perturbations and low for large perturbations in that range.

The verification setup is based on the 2025 VNN-COMP competition's setup and results \cite{kaulen20256th}. The competition report shows that the vast majority of instances solved by $\alpha, \beta$-CROWN are done within 5 minutes, and the number of instances solved plateaus after 5 minutes (Figure 5, page 27). We therefore set the verification time limit for each property to be 5 minutes.

Regarding hardware setup, we adopted a hardware configuration from VNN-COMP, namely the \textit{p3.2xlarge} instance on Amazon Web Services with 8 vCPUs, 61 GB of RAM, and a V100 GPU with 16 GB of memory. However, we found that the running script for $\alpha, \beta$-CROWN in VNN-COMP is configured to use only 1 CPU core, so we adjusted the CPU setup accordingly. In addition, since we do not have access to a V100 GPU, we decided to use 1/4 of the H100 SXM GPU's compute power with 20 GB of GPU memory, since these two setups have similar FP32 FLOPs (15.7 TFLOPs for V100 \cite{NVIDIATe6:online} and 16.75 TFLOPs for 1/4 of a H100 SXM GPU \cite{H100GPUN92:online}).

Our final verification hardware, therefore, is a single core of an AMD EPYC 9454 (Zen 4) CPU, 61 GB of RAM, and 1/4 of an NVIDIA H100 SXM5 GPU with 20 GB of memory.

\subsection{Generalization of Certified Robustness Across Test Sets}

Following the methodology of Jordan (2023) \cite{jordan2023variance}, we investigate whether ``lucky'' model seed in one test set can maintain its above-average certified robustness in another test set.

We split the 100 properties into two equal test sets, with the first 50 properties in one set and the remaining in the other, and evaluate how many properties each model satisfies in each test set. We calculate the correlation between the number of properties satisfied in the two test sets. A high correlation means better generalization of certified robustness across the test set and vice versa.

\section{Results \& Discussions}

We aim to answer the following research questions:

\begin{itemize}
    \item \textbf{RQ1:} What is the variance of certified robustness for models trained on different random seeds?
    \item \textbf{RQ2:} Does the standard deviation of certified robustness estimates exceed the marginal gains reported in recent machine learning literature?
    \item \textbf{RQ3:} Do ``lucky'' model seeds maintain their certified robustness in another test set?
\end{itemize}

We find that the standard deviation in certified robustness across different seeds is extremely large, and is statistically significantly larger than the marginal improvements reported in much of the machine learning literature on certified robustness. We also find that certified robustness may not generalize well across different test sets, since the correlation in certified robustness between two test sets is strong on MNIST but weak on Mars Frost Identification, with a large difference between the two datasets.

In the following, we report our results in more detail and discuss their implications.

\subsection{RQ1: Certified Robustness is Subjected to Extreme Variance Due to Random Seeding}

We summarize the neural network verification results in Table \ref{cr_variance}, which reports the mean and standard deviation of the number of properties successfully verified across the random seeds. We underline the highest standard deviation in certified robustness for each dataset.

\begin{table}
    \centering
    \begin{tabular}{cc|ccc}
        \hline
        \textbf{Dataset} & \textbf{Acc Stddev} & \textbf{$\epsilon$} & \textbf{CR Mean} & \textbf{CR Stddev}\\
        \hline
        \multirow{3}{*}{MNIST} & \multirow{3}{*}{0.057} & 0.006 & 81.6 & \textbf{22.3}\\
         && 0.007 & 54.3 & \textbf{\underline{28.9}}\\
         && 0.008 & 22.3 & \textbf{20.5}\\
        \hline
        \multirow{6}{*}{Mars Frost} & \multirow{6}{*}{0.198} & 0.0003 & 77.3 & \textbf{6.0}\\
         && 0.0005 & 67.5 & \textbf{4.9}\\
         && 0.0006 & 62.1 & \textbf{5.9}\\
         && 0.0007 & 55.8 & \textbf{6.3}\\
         && 0.0008 & 48.4 & \textbf{6.0}\\
         && 0.0009 & 39.1 & \textbf{\underline{11.8}}\\
        \hline\\
    \end{tabular}
    \caption{Standard deviation in accuracy and certified robustness of MNIST and Mars Frost models}
    \label{cr_variance}
\end{table}

We discover a surprising and extreme result: the certified robustness stddev is multiple orders of magnitude larger than the accuracy stddev. MNIST models with a 0.057\% accuracy stddev have 20.5\% - 28.9\% certified robustness stddev, representing a 360x - 500x increase. On the other hand, Mars Frost models with a 0.198\% accuracy stddev have 4.9\% - 11.8\% certified robustness, which is a 25x - 60x increase.

We believe our results have multiple concerning implications for dependable machine learning and neural network verification. Firstly, there will be serious doubts about the trustworthiness of certified robust models. One might argue that one can simply select the model seeds with the highest certified robustness, because the verification is for that specific model, not for the model architecture. However, we find many problems with this approach: (1) Doing so treats the random seeds as a tunable hyperparameter, and therefore is essentially performing post-validation hyperparameter tuning, which is not a sound methodology \cite{henderson2018deep} \cite{bouthillier2021accounting}. (2) Machine learning models ideally should achieve high performance (in terms of accuracy, robustness, etc.) by learning the structure of the underlying system \cite{d2022underspecification} \cite{geirhos2020shortcut} \cite{ilyas2019adversarial}. Achieving massive performance gains by changing the random seed (which is possible due to extreme variance) provides little confidence that models are genuinely learning the underlying structure.

Secondly, although there is variance in, e.g., model accuracy, Jordan (2023) \cite{jordan2023variance} offers a silver lining by showing that there is little variance in accuracy across the \textit{underlying test distributions}. However, we have major doubts that this result extends to certified robustness: Given that the variance in certified robustness is very high, we believe it is highly unlikely for models to also exhibit little variance with respect to the test distributions. In other words, while the selection of the test set from the underlying distribution can explain accuracy variance, it is unlikely to explain certified robustness variance. This makes it much harder to address certified robustness variance, since solutions such as adding more test data are likely to be ineffective.

\subsection{RQ2: Certified Robustness Stddev Exceeds Marginal Gains in Existing Works}

We surveyed papers on improving the certified robustness of neural networks, comprising 11 papers shown in Table \ref{paper_variance}, in which we record the largest marginal difference observed in any experiment. The highest and second-highest marginal improvements recorded are 9.85\% and 8.5\%, which are lower than the maximum stddev for both the MNIST and Mars Frost datasets.

Given that we are using a small number of seeds, we further perform a chi-square test to assess whether the population standard deviation of the models' certified robustness exceeds 9.85\% and 8.5\%. We present our results in Table \ref{chi_square}, where we find that the stddev of the certified robustness of Mars Frost models is statistically significantly larger than 8.5\%, although the result is not quite significant compared to 9.85\%. This, however, still means that the certified robustness stddev is statistically significantly larger than all marginal improvements we record except for one. On the other hand, for MNIST models, there is very strong statistical evidence that the stddev of their certified robustness exceeds even the largest marginal improvements we record.

This finding raises a crucial concern for the field of certified robust machine learning because it implies that marginal improvements in research are likely no larger than the stddev, making such results unconvincing. Importantly, since none of the papers we survey report uncertainty estimates or confidence intervals, the strength of these results is likely to have been overestimated. In the context of safety-critical applications, models being less dependable than previously thought is especially worrying, given the potential consequences in terms of money or human well-being.

\begin{table}
    \centering
    \begin{tabular}{cc|cc}
        \hline
        \textbf{Dataset} & \textbf{CR Stddev} & \textbf{CR Improvement} & \textbf{p-value}\\
        \hline
        \multirow{2}{*}{MNIST} & \multirow{2}{*}{28.9\% } & 9.85\% & \textbf{5.53e-13}\\
        && 8.5\% & \textbf{$\approx$ 0.0}\\
        \hline
        \multirow{2}{*}{Mars Frost} & \multirow{2}{*}{11.8\%} & 9.85\% & 0.171\\
        && 8.5\% & \textbf{0.0456} \\
        \hline
        \\
    \end{tabular}
    \caption{Chi-square test of whether stddev of models' certified robustness exceeds marginal improvements in existing works. Statistically significant results are \textbf{bolded}.}
    \label{chi_square}
\end{table}

\subsection{RQ3: ``Lucky'' Model Seeds Cannot Guarantee Generalization to Another Test Set}

We evaluate the models' certified robustness on two test sets and calculate their correlations to investigate whether ``lucky'' model seeds maintain their performance in another test set.

\begin{table}
    \centering
    \begin{tabular}{ccc}
        \hline
        \textbf{Dataset} & \textbf{$\epsilon$} & \textbf{Correlation}\\
        \hline
        \multirow{3}{*}{MNIST} & 0.006 & 98.9\%\\
         & 0.007 & 98.6\% \\
         & 0.008 & \textbf{96.6\%} \\
        \hline
        \multirow{6}{*}{Mars Frost} & 0.0003 & 83.4\%\\
         & 0.0005 & 41.6\% \\
         & 0.0006 & \textbf{31.1\%} \\
         & 0.0007 & 45.5\% \\
         & 0.0008 & 57.6\% \\
         & 0.0009 & 73.3\% \\
        \hline\\
    \end{tabular}
    \caption{Correlation between two test sets of certified robustness: ``Lucky'' model seed generalization is strong for mnist but weak for mars frost}
    \label{cr_generalization}
\end{table}

\begin{figure}[h]
    \centering
    \includegraphics[width=0.7\linewidth]{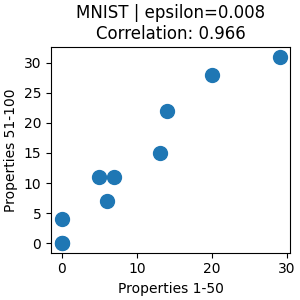}
    \caption{\textbf{Correlation of certified robustness between test sets for MNIST models at perturbation $\epsilon = 0.008$.} A strong correlation is observed, showing that certified robustness generalizes well for MNIST models.}
    \label{mnist_corr}
\end{figure}

\begin{figure}[h]
    \centering
    \includegraphics[width=0.7\linewidth]{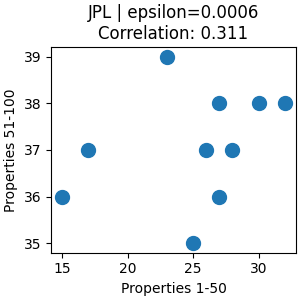}
    \caption{\textbf{Correlation of certified robustness between test sets for Mars Frost models at perturbation $\epsilon = 0.0006$.} A weak correlation is observed, showing that certified robustness does \textbf{not} generalize well for Mars Frost models.}
    \label{jpl_corr}
\end{figure}

We report our results in Table \ref{cr_generalization}. We find that the lucky MNIST model seeds generalize very well to a new test set, with the lowest correlation being 96.6\% across the tested perturbations. However, the lucky Mars Frost model seeds generalize much worse, with the lowest correlation of only 31.1\%. We show the exact data points for these two cases in Figure \ref{mnist_corr} for MNIST models and Figure \ref{jpl_corr} for Mars Frost models.

Our results, therefore, demonstrate that random seed generalization varies widely across datasets; hence, it cannot be guaranteed that lucky model seeds will also be certified robust when evaluated on a second test set. A major implication of this result, as observed in \cite{jordan2023variance}, is that selecting the best model seed post-verification is not only unsound, as previously discussed, but may also fail to yield a superior model that generalizes well to unseen data.

\section{Conclusion}

Our paper examines the extreme extent and practical consequences of variance in certified robustness of neural networks. We demonstrate that such variances are surprisingly large, significantly larger than the accuracy variance, and statistically larger than most marginal improvements in certified robustness in existing works. We also show that model seeds with better-than-average certified robustness may not maintain their performance in a second test set.

In light of these results, we once again urge researchers working on certified robustness to report uncertainty quantification or confidence intervals to be transparent about whether their models can be trusted in safety-critical applications. For those verifying neural networks, we urge them to use large-scale, diverse, and unseen datasets, including those not directly used for training, to more comprehensively evaluate whether the certified robustness of verified models is genuinely generalizable to unseen data.

\bibliographystyle{IEEEtran}
\bibliography{references}

\end{document}